\documentclass{article}

\usepackage{color,soul}

\usepackage{multirow}

\usepackage{amssymb}
\usepackage{mathtools}
\usepackage{bbm}

\usepackage{microtype}
\usepackage{graphicx}
\usepackage{subfigure}
\usepackage[width=.75\textwidth]{caption}
\usepackage{booktabs} 

\usepackage{amsfonts}
\usepackage{amssymb}

\usepackage{natbib}

\usepackage{todonotes}

\newcommand{\etal}{{\em et al.}}

\newcommand{\algoname}{{\sc FPF}}
\newcommand{\rf}{{\sc RF}}
\newcommand{\rsm}{{\sc RSM}}

\newcommand{\dbc}{{\sc Breast\_Cancer}}
\newcommand{\dwn}{{\sc Wine}}
\newcommand{\dsb}{{\sc Spam\_Base}}

\newcommand{\dataset}{\ensuremath{{\cal D}}}
\newcommand{\forest}{\ensuremath{{\cal T}}}
\newcommand{\inputspace}{\ensuremath{{\cal X}}}
\newcommand{\feats}{\ensuremath{{\cal F}}}
\newcommand{\labels}{\mathcal{Y}}

\newcommand{\R}{\mathbb{R}}

\newcommand{\leaf}[1]{\lambda(#1)}
\newcommand{\node}[1]{\sigma(#1)}

\renewcommand\vec{\boldsymbol}

\usepackage[accepted]{icml2019}

\icmltitlerunning{Feature Partitioning for Robust Tree Ensembles and their Certification in Adversarial Scenarios}

\newtheorem{definition}{Definition}

\newtheorem{proposition}{Proposition}

\begin{document}

\twocolumn[
\icmltitle{Feature Partitioning for Robust Tree Ensembles\\ and their Certification in Adversarial Scenarios}

\icmlsetsymbol{equal}{*}

\begin{icmlauthorlist}
\icmlauthor{Stefano Calzavara}{unive,equal}
\icmlauthor{Claudio Lucchese}{unive,equal}
\icmlauthor{Federico Marcuzzi}{unive,equal}
\icmlauthor{Salvatore Orlando}{unive,equal}
\end{icmlauthorlist}

\icmlaffiliation{unive}{Department of Environmental Sciences, Informatics and Statistics, Ca' Foscari University of Venice, Italy}

\icmlcorrespondingauthor{Claudio Lucchese}{claudio.lucchese@unive.it}

\icmlkeywords{Adversarial Machine Learning}

\vskip 0.3in
]




\printAffiliationsAndNotice{\icmlEqualContribution} 

\begin{abstract}
Machine learning algorithms, however effective, are known to be
vulnerable in adversarial scenarios where a malicious user may
inject manipulated instances.
In this work we focus on evasion attacks, where a model is trained in a safe environment and exposed to attacks at test time.
The attacker aims at finding a minimal perturbation of a test instance that changes the model outcome.

We propose a model-agnostic strategy that builds a robust ensemble by training
its basic models on feature-based partitions of the given dataset.
Our algorithm guarantees that the majority of the models in the ensemble cannot be affected by the attacker. We experimented the proposed strategy on decision tree ensembles, and we also propose an approximate certification method for tree ensembles that efficiently assess the minimal accuracy of a forest on a given dataset avoiding the costly computation of evasion attacks.

Experimental evaluation on publicly available datasets shows that proposed strategy outperforms state-of-the-art adversarial learning algorithms against evasion attacks.

\end{abstract}


\section{Introduction}
\label{sec:intro}

Machine Learning (ML) algorithms are currently used to train models that are then deployed to ensure system security and to control critical processes~\cite{HuangJNRT11,BiggioR17}. Unfortunately, traditional ML algorithms proved vulnerable to a wide range of attacks, and in particular to \emph{evasion attacks}, where an attacker carefully craft perturbations of an inputs to force prediction errors~\cite{BiggioCMNSLGR13,NguyenYC15,PapernotMJFCS16,Moosavi-Dezfooli16}. 

While there is a large body of research on evasion attacks in linear classifiers~\cite{LowdM05,BiggioNL11} and, more recently, on deep neural networks~\cite{SzegedyZSBEGF13,GoodfellowSS15}, there are a few works dealing with tree-based models. Decision trees are \emph{interpretable} models~\cite{tolomei2017kdd}, yielding predictions which are human-understandable in terms of syntactic checks over domain features, which is particularly appealing in the security setting. Moreover, \emph{decision trees ensembles} are nowadays one of the best methods for dealing with non-perceptual problems, and are one of the most commonly used techniques in
Kaggle competitions~\cite{Chollet:2017:DLP:3203489}. 

In this paper we present an algorithm, called \emph{Feature Partitioned Forest} (\algoname), that builds an ensemble of decision trees aimed to be robust against evasion attacks. Indeed, the trained ensemble is a binary classifier that we show to be,  in most of the cases, \emph{robust by construction}. In fact, given a test instance that the attacker aims to corrupt, if most of the trees in the ensemble returns accurate binary predictions for that instance, the attacker has no chance to attack the whole ensemble. Our method is based on a particular sampling of the features, where we randomly equi-partition the set of features, and train each tree on a distinct feature partition. Moreover, as usual in the context of adversarial learning, we count on a threat model that limits the budget of an attacker, that can only manipulate a limited number of features, thus upper bounding the distance between the original instance and the perturbed one. We also propose an approximate \emph{certification method} for our tree ensembles that efficiently assesses the minimal accuracy of a forest on a given dataset, avoiding the costly computation of evasion attacks.


\section{Related Work}
\label{sec:related}

Most of the work in this adversarial learning regards classifiers, in particular binary ones. The attacker starts from a positive instance that is classified correctly by the deployed ML model and is interested in introducing minimal perturbations on the instance to modify the prediction from positive to negative, thus ``evading'' the classifier~\cite{NelsonRHJLLRTT10,BiggioCMNSLGR13,BiggioFR14,SrndicL14,KantchelianTJ16,Carlini017,DangHC17,GoodfellowSS15}. To prevent these attacks, different techniques have been proposed for different models, including support vector machines~\cite{BiggioNL11,XiaoBNXER15}, deep neural networks~\cite{GuR14,GoodfellowSS15,PapernotM0JS16}, and decision tree ensembles~\cite{KantchelianTJ16,ChenZBH19}. Unfortunately, the state of the art for decision tree ensembles is far from satisfactory. 

The first adversarial learning technique for decision tree ensembles is due to Kantchelian {\em et al.} and is called \emph{adversarial boosting}~\cite{KantchelianTJ16}. It is an empirical data augmentation technique, 
borrowing from the {\em adversarial training} approach \cite{SzegedyZSBEGF13}.
Another adversarial learning technique for decision tree ensembles was proposed in a recent work by Chen {\em et al.}, who introduced the first tree learning algorithm embedding the attacker directly in the optimization problem solved upon tree construction~\cite{ChenZBH19}. 
The key idea of their approach, called \emph{robust trees}, is to redefine the splitting strategy of the training examples at a tree node. 
Finally, our algorithm \algoname{} has some relations with Random Subspace method (\rsm) \cite{ho1998random}, which was successfully exploited by \citealp{biggio2010multiple} to build ensembles where each single model is trained on a projection of the original dataset on a subset of its features.

\section{Robust Forest Training}
\label{sec:alg}

We aim to design a machine learning algorithm able to train forests of binary decision trees that are resilient to \emph{evasion attacks}, which occurs when an adversary adaptively manipulates test data to force prediction errors.
Specifically, in this section we discuss our algorithm able to train a robust forest that is resilient in a strong adversarial environment, where attackers can perturb at most $b$ features of a test instance to deceive the learnt model and force a prediction error.  We first introduce some notation and the threat model, then we discuss our algorithm.

\subsection{Background and Notation}
Let $\inputspace \subseteq \R^d$ be a $d$-dimensional vector space of real-valued \textit{features}. An {\em instance} $\vec{x} \in \inputspace$ is a $d$-dimensional feature vector $(x_1, x_2, \ldots, x_d)$, where we denote with $\feats$ the set of features.
Each instance $\vec{x} \in \inputspace$ is assigned a label $y \in \labels$ by some unknown \emph{target} function $g: \inputspace \mapsto \labels$.  The goal of a \emph{supervised learning} algorithm that induces a \textit{forest of decision trees} is to find the forest $\forest$ that best approximates the target $g$.

In this paper we discuss binary classification, where $\labels = \{-1,+1\}$, and focus on binary decision trees.  
Each tree $t \in \forest$ can be inductively defined as follows: $t$ is either a leaf $\leaf{\hat{y}}$ for some label $\hat{y} \in \labels$, or a internal test node $\node{f,v,t_l,t_r}$, where $f \in [1,d]$ identifies a feature, $v \in \R$ is the threshold for the feature $f$, and $t_l,t_r$ are left/right decision trees. 

At test time, an instance $\vec{x}$ traverses each tree $t \in \forest$ until it reaches a leaf $\leaf{\hat{y}}$, which returns the \textit{prediction} $\hat{y}$, denoted by $t(\vec{x}) =  \hat{y}$. Specifically, for each internal test node $\node{f,v,t_l,t_r}$, $\vec{x}$ falls into the left tree $t_l$ if $x_f \leq v$, and into the right tree $t_r$ otherwise.  
Given a forest \forest, the global prediction 
is defined as $\forest(\vec{x})=+1$ 
if $\sum_{t \in \forest} t(\vec{x})>0$, 
and $\forest(\vec{x})=-1$ otherwise.

Finally, given a test set $\dataset_{test}$, let $E = \{(\vec{x}, y) \in  \dataset_{test} \; | \; y \cdot \forest(\vec{x}) < 0\}$ be the set of test instances that are not classified correctly by $\forest$.
We can finally define the \emph{accuracy} as:
\begin{equation}
\label{eq:acc}
\nonumber
Acc = \frac{|\dataset_{test}| - |E|}{|\dataset_{test}|}
\end{equation}

\subsection{Threat Model}

We focus on the {\em evasion attack} scenario, where an attacker aims at fooling an already trained classifier by maliciously modifying a given instance before submitting it to the classification model. The perturbation caused by the attacker is not unconstrained as the attack should be ``invisible'' to the classification system.

As in Kantchelian \etal~\cite{kantchelian2016evasion}, we assume an attacker $A_b$ that is capable of modifying a given instance $\vec{x}$
into a perturbed instance $\vec{x}'$ such that the $L_0$-norm of the perturbation is smaller than the attacker's budget $b$, i.e., $\lVert \vec{x} - \vec{x}' \rVert_0 \leq b$. 
Therefore, attacker $A_b$ can perturb the instance $\vec{x}$ by modifying at most $b$ features, without any constraint on how much a given feature can be altered.
Indeed, a very small $b$ is sufficient to achieve successful attacks.
Su \etal~\cite{su2019one} show that with a one-pixel attack, i.e., with $b=1$, it is possible to fool a complex deep neural network as VGG16~\cite{simonyan2014very} and decrease its accuracy to a poor 16\%.

Given an instance $\vec{x}\in\inputspace$, we denote by $A_b(\vec{x})$ the set of all the perturbed instances the attacker may generate:
$$
A_b(\vec{x}) = 
\left\{ \vec{x}' ~|~ \vec{x}'\in\inputspace\ \wedge\ \lVert \vec{x} - \vec{x}' \rVert_0 \leq b
\right\}.
$$

Finally, we can define the \emph{accuracy under attack} that an attacker $A_b$ aims to \emph{minimize}. 
Given the test set $\dataset_{test}$, let $E = \{(\vec{x}, y) \in  \dataset_{test} \; | \; y \cdot \forest(\vec{x}) < 0\}$. We define $E' = \{(\vec{x}, y) \in \dataset_{test} \setminus E \; | \; \exists \vec{x}' \in A_b(\vec{x}), \ y \cdot \forest(\vec{x}') < 0\}$ be the set of the perturbed instances that are not classified correctly by $\forest$. 
We can finally define the accuracy under attack as:
\begin{equation}
\label{eq:accua}
Acc_{A_b} = \frac{|\dataset_{test}| - |E \cup E'|}{|\dataset_{test}|}
\end{equation}


\subsection{Our algorithm}

In the following we propose a novel training strategy that produces a forest $\forest$ where the majority of its trees are not affected by the attacker $A_b$.

\paragraph{Feature Partitioning.} Given a \emph{set partition} $\mathcal{P}$ of the feature set $\feats$ and an attacker $A_b$
that decided to corrupt the set of features $B\subseteq \feats$, with $|B|\leq b$,
we can easily compute the number of sets in $\mathcal{P}$ overlapping with $B$ as:\footnote{$\mathbbm{1}{[e]}$ equals $1$ if expression $e$ is true and $0$ otherwise.}
$$
O(\mathcal{P},B) = \sum_{P \in \mathcal{P}} \mathbbm{1}{[B \cap P \neq \emptyset]}.
$$

We call $\mathcal{P}$ {\em robust} if the majority of its sets cannot be impacted by the attacker $A_b$, i.e., if the following property holds:
$$
\forall B\subseteq\feats, |B|\leq b \quad O(\mathcal{P},B) < 
\frac{\lvert \mathcal{P} \rvert}{2}.
$$




When $|B|\leq b$, it is straightforward to show that this property is surely satisfied if  $\lvert\mathcal{P}\rvert \geq 2b+1$. Consider the worst case: at most $b$ distinct subsets of $\mathcal{P}$ can have an overlap with $B$, leaving other $\geq b+1$ subset of $\mathcal{P}$ unaffected. Hereinafter, we consider only {\em robust} feature partitions $\mathcal{P}$ where $|\mathcal{P}|=2b+1$. 

\paragraph{Robust forest.} 

Let's consider a forest $\forest$ that, given an attacker $A_b$, is built by exploiting a robust feature partition $\mathcal{P}$ as follows.

Let $\dataset$ be a set of training instances $\vec{x}\in\inputspace$,
and $\mathcal{P}$ be a robust partition of its feature space $\feats$.
Given $P \in \mathcal{P}$, we call $\pi_{P}(\dataset)$ the projection of $\dataset$ on the feature set $P$, i.e., the dataset obtained from $\dataset$ by discarding those features not included in $P$.
Given a robust feature partitioning ${\cal P}$, it is thus possible to build a robust forest by training $2b+1$ trees independently on the $2b+1$ projections $\pi_{P}(\dataset)$, with $P\in {\cal P}$. 

The algorithm sketched above achieves what we formally define as {\em robustness}.

\begin{definition}[Robust Forest]
\label{def:robustness}
Given an attacker $A_b$, we say that a forest ${\cal T}$ is {\em robust} if the majority of its trees is not affected by $A_b$ for any of its attacks:
$$
\forall \vec{x}\in\inputspace, \forall\vec{x}'\in A_b(\vec{x}) \quad
 \sum_{t \in \forest} \left(\mathbbm{1}{[t(\vec{x})=t(\vec{x}')]}\right)>
 \frac{\lvert \forest \rvert}{2}
$$
\end{definition}

It is straightforward to show that if a forest is built on the basis of a robust feature partitioning $\mathcal{P}$ as described above,
then, at most $b$ of its $2b+1$ trees can be affected by the attacker.

In the best case scenario where each $t \in \forest$ is perfectly accurate,
the above robustness property provides that, in presence of attacks, 
only a minority of trees provides an incorrect prediction. 
Clearly, this scenario is unlikely, and therefore we discuss below how to strengthen the accuracy of $\forest$.

Note that, the above definition and training strategy trivially generalizes to any ensemble learning algorithm.

\paragraph{Increasing the accuracy of a robust forest.} 
The above definition does not provide any guarantee on the accuracy of the full forest $\forest$, which clearly depends on the accuracy of its single trees. Yet, the more accurate the trees $t \in \forest$, the more likely the forest $\forest$ is accurate under attack.

The accuracy of single trees depends on the feature partitioning $\mathcal{P}$. The larger $|\mathcal{P}|$ the smaller is the number of features each tree can be trained on. To increase the accuracy of a robust forest $\forest$, we equi-partition $\feats$ across  $\mathcal{P}$ so as to have $|P| \geq \lfloor\frac{\lvert\feats\rvert}{2b+1}\rfloor$ for all $P \in \mathcal{P}$. Clearly, as the attacker's power $b$ increases, we require to partition $\feats$ in to a larger number of subsets, and for these to be effective we need the dataset to have a larger number of high quality features. Note that this is true for every learning algorithm: if the attacker can perturb at will up to $b$ features, it is necessary to have more than $b$ high quality features to train an accurate model.

In addition, a specific partitioning $\mathcal{P}$ may only be sub-optimal
as there may be multiple way of partitioning $\feats$ so as to achieve features subsets with high predictive power. Along the lines of ensemble training, we use multiple feature partitionings and join together the resulting robust forests in a single decision tree ensemble.

\begin{algorithm}[tb]
   \caption{\algoname}
   \label{alg:myalgo}
\begin{algorithmic}
   \STATE {\bfseries Input:} dataset $\dataset$, attacker $A_b$, 
   training rounds $r$
   \STATE $\forest \gets \emptyset$
   \STATE $k \gets 2 * b + 1$
   \FOR{$i=1$ {\bfseries to} $r$}
     \REPEAT
     \STATE $\forest_i \gets \emptyset$
     \STATE $\mathcal{P}_i \gets {\sf RandomPartition}(\feats, k)$\label{alg:rndpart}
     \FORALL{$P_j \in \mathcal{P}_i,\  j=1,\ldots,k$}
       \STATE $t \gets {\sf DecisionTree}(\pi_{P_j}(\dataset))$
         \STATE $\forest_i  \gets \forest_i \cup \{t\}$
     \ENDFOR
     \UNTIL{${\sf AcceptCondition}(\forest_i)$}
   \STATE $\forest \gets \forest  \cup \forest_i $
   \ENDFOR
   \STATE {\bf return} $\forest$.
\end{algorithmic}
\end{algorithm}

Finally, we can sketch our algorithm \algoname{} to train a forest $\forest$ aimed to be robust against an attacker $A_b$ that can perturb at most $b$ features. 
The algorithm, shown in Alg.~\ref{alg:myalgo}, iterates a number $r$ of user-defined rounds. During each round $i$, the algorithm generates a random feature partitioning 
$\mathcal{P}_i$ of the features $\feats$ present in the given training dataset $\dataset$. The feature set $\feats$ is randomly and evenly split into $k=2b+1$ disjoint subsets, and a new decision tree is trained on each of the dataset projections $\pi_{P}(\dataset)$ for every feature subset $P \in {\cal P}_i$.
The resulting $2b+i$ trees form a tree ensemble $\forest_i$.
We use an accept condition to filter out those $\forest_i$ that would not strengthen the final forest. For instance, it might be the case that the some partitions in $\mathcal{P}_i$ do not contain sufficiently predictive features to train accurate trees. In this work, we use a simple acceptance criteria according to which a $\forest_i$ is accepted if its accuracy is larger than na\"ivly predicting the dataset's majority class. In this case, the trees of $\forest_i$ are added to the forest $\forest$.
Eventually the returned forest contains a total of $r (2b+1)$ trees.

\begin{proposition}[Robustness of \algoname]
The forest $\forest$ built by the algorithm \algoname{} is {\em robust}
against and attacker $A_b$ as the majority of its tree is not affected by $A_b$ for any of its attacks.
\end{proposition}

At each round \algoname{} trains a set of $2b+1$ trees, which, as discussed above,
is robust because at most $b$ of its trees can be affected by $A_b$.
After the same reasoning applied to $r$ rounds, \algoname{} builds a forest $\forest$
of $r(2b+1)$ trees of which at most $rb$ can be affected by $A_b$,
leaving a majority of $r(b+1)$ trees unaltered.

\section{Evaluation and certification of tree-based models}

Evaluating the accuracy of a model in presence of an attacker is
a difficult and computationally expensive task.
This is due to the possibly large size of $A_b(\vec{x})$ for some $\vec{x}\in\inputspace$
and to the number of interactions among trees in a forest.
\citealp{chen2019robustness} show that verifying the robustness of a forest $\forest$ with at most $l$ leaves per tree has cost
$\min\{O(l^{|\forest|}), O((2|\forest|l)^{|\feats|})\}$ assuming a $L_\infty$-norm attacker. \citealp{kantchelian2016evasion} prove that in case of a $L_0$-norm attacker, as in this work, the problem of finding a successful attack is NP-complete.

Below we first provide an expensive brute-force strategy for evaluating the accuracy under $L_0$-norm attack, then we show that the evaluation problem can be reduced to a maximum coverage problem and we propose a few very efficient heuristic strategies that can be used both to reduce the cost of the brute-force strategy and to provide a lower-bound certification for a tree-ensemble model on a given dataset.

\subsection{Brute-force evaluation}

Given an instance $\vec{x} \in \inputspace$, the brute-force evaluation of a forest $\forest$ consists in generating all the possible perturbations an attacker $A_b$ is capable of to find whether there exists $\vec{x'}\in A_b(\vec{x})$ such that 
$\mathcal{T}(\vec{x}) \neq \mathcal{T}(\vec{x'})$.

The size of $A_b(\vec{x})$ is infinite, but we can limit its enumeration to the
set of attacks that are relevant for the given forest $\forest$, i.e., those attacks
that can invert the outcome of a test in some internal nodes of trees in $\forest$.
Recall that nodes in a tree are in the form $x_f\leq v$ for some threshold $v$.
Indeed, the thresholds used in the tree nodes induce a discretization of the input space $\inputspace$ that we exploit as follows.

For any given feature $f\in\feats$, we define with ${\cal V}_f$ the set of relevant thresholds as follows:
$$
{\cal V}_f = \{ v ~|~ \exists \node{f,v,t_l,t_r} \in t, t \in \forest\} \cup \{\infty\}
$$

The set ${\cal V}_f$ includes all the thresholds
that are associated to $f$ in any node $\node{f,v,t_l,t_r}$ of any tree in $\forest$,
plus the $\infty$ value that allows to traverse the right branch of the node with the largest threshold.

Given an attacker $A_b$, the set of relevant perturbations is thus given by the
cartesian product of sets ${\cal V}_f$ for $b$ different features. Let $\feats_b$ be the set of all subsets $F\subseteq\feats$ having size at most $b$, we denote with $\hat{A}_b(\vec{x})$ the set of such perturbations, formally defined as:
$$
\hat{A}_b(\vec{x}) = \left\{\vec{x'} ~|~  x_f = v, \forall v \in \mathcal{V}_f, \forall f \in F, F\in \feats_b\right\}
$$

We conclude that an attacker $A_b$ can successfully perturb an instance $\vec{x}$
against a forest $\forest$ if there exists at least one $\vec{x'} \in \hat{A}_b(\vec{x}) $ for which $\mathcal{T}(\vec{x}) \neq \mathcal{T}(\vec{x'})$.

This brute-force approach is very expensive, due to three factors: {\em i)} as $b$ increases, the number of feature combinations $\feats_b$ increases; {\em ii)} as the number of trees and nodes grows, the number of threshold values associated with each feature increases; {\em iii)} for each perturbed instance $\vec{x'}$, the prediction $\mathcal{T}(\vec{x'})$ must be retrieved by traversing the given forest.

\subsection{Attacking forest $\forest$ as a Maximum Coverage Problem.}

Given a forest $\forest$ and an input instance $\vec{x}$
the attacker $A_b$ aims at finding those perturbations that
lead the majority of the trees to a wrong prediction.
Indeed, as some trees of  $\forest$ might give incorrect predictions before
the attack, it could be sufficient to harm less than $\lceil\forest/2\rceil$ trees.

We now introduce some simplifying assumptions and then show that finding 
an attack can be reduced the the maximum coverage problem.
First, we assume that if a tree in $\forest$ provides a wrong prediction before the attack, then its prediction will be incorrect also after the attack.
Second, we assume that if a tree uses a feature $f$ for its prediction over $\vec{x}$ then attacking $f$ causes the tree to generate a wrong prediction.

Note that these assumptions are very conservative. An incorrect tree may, by chance, provide a good prediction after the attack. More importantly, modifying a feature $f$ does not necessarily flips the test performed on every node using that feature and leads to a wrong prediction. These assumptions allow to clear formulation of the problem, and our experiments show that the error introduced is interestingly small.

Under the above assumptions, the aim of the attacker $A_b$ is to find a set of $b$ features that are used by the largest number of distinct trees.
Let's denote with $S_f$ the set of trees in $\forest$ using feature $f$,
and let $S$ be $S=\bigcup_{f\in\feats} S_f$.
Then the most successful attack is given by the subset $S'\subseteq S$ with $|S'|\leq b$ such that $|\cup_{S_i \in S'}|$ is maximized. 
The thoughtful reader has surely recognized that this formulation of our problem is nothing else that an instance of the \emph{maximum coverage problem}.

Note that algorithm \algoname{} limits the use of a single feature to at most $r$ trees (the number of rounds) out of a total of $r(2b+1)$ trees, to some extent, making it more difficult for the attacker to find a cover.

Before attacking the maximum coverage problem we make a few improvements
to provide a more accurate definition of sets $S_f$.

First, we do not consider trees in $\forest$ with an incorrect prediction before the attack.

Second, we note that a tree may include a feature $f$ in some of its nodes,
but these nodes may never be traversed during the evaluation of an instance $\vec{x}$.
Therefore we say that a tree $t$ belongs to $S_f$ for an instance $\vec{x}$
only if the traversal path of $\vec{x}$ in $t$ includes a node with a test on feature $f$. 

Last, among the nodes along the traversal path of instance $\vec{x}$ before the attack, we distinguish between nodes where the test $x_f\leq v$ is true, and nodes where the test is false. In the former case, the attacker must increase the value of $x_f$ to affect the traversal path, while in the latter case $x_f$ should be decreased. Clearly, these two attacks cannot coexist. Therefore, we
define sets $S_f^+$ and $S_f^-$, where we include a tree $t$ in $S_f^+$
if feature $f$ is in the traversal path of $\vec{x}$ with a true outcome of the test $x_f\leq v$, and in $S_f^-$ otherwise.

We thus achieved a more accurate modeling of when an attack can actually affect the final prediction. This also reduces size of sets $S_f^+$, $S_f^-$ decreasing the risk of overestimating the effect of an attack.

We can finally summarize the maximum coverage problem as follows.
Given the set $S=\bigcup_{f\in\feats} S_f^+ \cup S_f^-$,
the most successful attack is given by the subset $S'\subseteq S$ with $|S'|\leq b$ 
under the constraint that if $S_f^+$ and $S_f^-$ cannot be included together in $S'$
that maximizes the cover of the (correct) trees in the forest $\forest$.

We say that there is no possible attack if the number of trees in the largest cover plus the number of trees providing a wrong prediction before attack is the majority.

Note that however, this is a conservative estimate as the attacker my modify all the features identified by the maximum cover without being able to affect the final forest prediction. In the following, we use this conservative set cover formulation to define heuristic strategies that can be used to provide a lower-bound to the accuracy of a forest on a given dataset, or to speed-up the brute-force approach by discarding those instances for which a sufficiently large cover does not exist.

\subsection{Fast Accuracy Lower Bound}

Given an attacker $A_b$, a forest $\forest$ and an instance $\vec{x}$,
we denote with $\omega$ the number of trees providing a wrong prediction
and with $S$ the elements of the set cover formulation as defined above.

It easy to provide an upper bound to the size of the largest cover as follows.
First sort the sets in $S$ according to their size. Then, we select the $b$ largest sets
by enforcing the constraint that $S_f^+$ and $S_f^-$ cannot be considered together.
Let $S_{FLB}$ be the covering sets selected as above, we know that the size of the largest cover cannot be larger than $\overline{S_{FLB}} = \sum_{S_i\in S_{FLB}} |S_i|$.
Therefore, we can pessimistically estimate the number of incorrect trees under attack as $\omega + \overline{S_{FLB}}$, which leads to an incorrect prediction over $\vec{x}$ if larger than $|\forest|/2$.

By applying the same algorithm to every correctly classified instance $\vec{x}\in\dataset$,
we denote with $E_{FLB}$ the set of instances for which $\omega+ \overline{S_{FLB}}\geq |\forest|/2$, obtaining a lower bound on the accuracy of the forest $\forest$ on dataset $\dataset$:

\begin{equation}
\nonumber
FLB(A_b,\dataset) = \frac{|\dataset| - |E \cup E_{FLB}| }{|\dataset|}.
\end{equation}





\subsection{Exhaustive Accuracy Lower Bound}

In order to improve over the fast lower bound, we also consider a more expensive option where all the possible covers are considered and the maximal is eventually found.

Given an attacker $A_b$, let $S$ the elements of the set cover formulation as previously introduced. The exhaustive search consists in enumerating all the possible
subset $S'\subseteq S$ of size at most $b$, and then for each of them
we compute the corresponding cover $|\bigcup_{S_i\in S'} S_i|$.
Let $\overline{S_{ELB}}$ be the maximum of such covers, 
we can define $E_{ELB}$ the set of instances in $\dataset$
for which $\omega+ \overline{S_{ELB}}\geq |\forest|/2$,
and introduce following accuracy lower bound:

\begin{equation}
\nonumber
ELB(A_b,\dataset) = \frac{|\dataset| - |E \cup E_{FLB}| }{|\dataset|}.
\end{equation}

\subsection{Reducing the cost of the brute-force evaluation}

Above we discussed some strategies to check whether there exists a cover
that identifies an harmful attack. Strategies have different costs: {\sc FLB}
requires to sort the candidate sets of the cover, while {\sc ELB} performs an exhaustive search. Both are however much cheaper than brute-force evaluation.

When the lower-bound information is not considered sufficient, we propose to exploit
the above strategies in the following way. Given an instance $\vec{x}$ and an attacker $A_b$, we proceed as follows:
\begin{enumerate}
\item first compute $\overline{S_{FLB}}$: if the cover is not sufficiently large, then the instance cannot be attacked; otherwise
\item compute $\overline{S_{ELB}}$: if the cover is not sufficiently large, then the instance cannot be attacked; otherwise
\item use the brute-force method to check the existence of a successful attack.
\end{enumerate}

Experimental results show that the above cascading strategy is able to strongly reduce the number of instances in a given dataset $\dataset$ for which the brute-force approach is required.

\section{Experiments}
\label{sec:exp}

\subsection{Experimental settings}

\begin{table}[t]
\caption{Datasets description.}
\label{tab:dataset}
\scriptsize
\vskip 0.15in
\begin{center}
\begin{sc}
\begin{tabular}{lcccr}
\toprule
Dataset & $|\feats|$ & \textnormal{\#top} $\feats$ & \textnormal{maj. class} & $|\dataset|$ \\
\midrule
\dwn{} & 13 & 7 & 73.0\% & 178 \\ 
\dbc{} & 30 & 15 & 62.7\% & 569 \\ 
\dsb{} & 57 & 26 & 60.6\% & 4600 \\ 
\midrule
MNIST 0/1 & 784 & 54 & 53.3\% & 14780 \\ 
MNIST 5/6 & 784 & 173 & 52.1\% & 13189 \\ 
MNIST 1/7 & 784 & 79 & 51.9\% & 15170 \\ 
\bottomrule
\end{tabular}
\end{sc}
\end{center}
\vskip -0.1in
\end{table}

In Table~\ref{tab:dataset} we report the main characteristics of the datasets used
in the experimental evaluation, including the number of features, the number of top relevant features measured as those contributing to the 90\% of the feature importance in a Random Forest, and the relative size of the majority class. Datasets, ranging from small to mid-sized, are associated to a binary classification task,
and they are commonly used in adversarial learning literature.\footnote{All dataset are available at the UCI Machine Learning Repository.} 

We compared our proposed algorithm \algoname{} against the following tree-ensemble competitors:
\begin{itemize}
	\item Random Forest (\rf) \cite{breiman2001randomforest}, which is known to have some good level of robustness thanks to the ensembling of several decision trees. As in the original algorithm, each tree is trained on a bootstrap sample of the dataset, with no constraints on the number of leaves, and with feature sampling of size $\sqrt{|\mathcal{F}|}$ at each node.
	\item Random Subspace method (\rsm) \cite{ho1998random} which was successfully exploited by \citealp{biggio2010multiple}. In this case, each tree is trained on a projection of the original dataset on a subset of its features. Validation experiments showed best results with 20\% feature sampling.
\end{itemize}

Hyper-parameter tuning on a validation set showed that both \algoname{} and \rsm{} perform best when limiting the number of leaves to $8$. Similarly, we limited the number of trees to $100$ for datasets \dbc{} and \dsb{}, and to $300$ for dataset \dwn{}. We observed that \dwn{} requires a larger forest due to its limited number of features. All results were computed on a randomly selected test set sized 1/3 of the original dataset. Hereinafter, we use $b$ to address the training parameter of \algoname, and we use $k$ for the attack strength of attacker $A_k$.

The following experimental evaluation aims at answering the following research questions:
\begin{itemize}
	\item is \algoname{} able to train more robust model?
	\item how is \algoname{} affected by the number of rounds $r$ and the expected attacker power $b$?
	\item how accurate are the proposed bound, and can we exploit them to efficiently analise models on larger attacker budgets $k$?
\end{itemize}

\subsection{Robustness Analysis}

\begin{table}[t]
\caption{Accuracy on the \dbc{} dataset.}
\label{tb:bc_comp}
\scriptsize
\vskip 0.15in
\begin{center}
\begin{sc}
\begin{tabular}{lcrrccc}
\toprule
\multirow{2}{*}{\textnormal{model}} & \multicolumn{3}{c}{\textnormal{parameters}} & \multicolumn{3}{c}{\textnormal{attacker}} \\
 & $b$ & $r$ & $|\forest|$ & $A_0$ & $A_1$ & $A_2$ \\
\midrule
\rf   & & & 100 & 0.96 & 0.91 & 0.80 \\
\rsm  & & & 100 & 0.96 & 0.92 & 0.87 \\
\midrule
\multirow{5}{*}{\algoname} 
 & 1 & 33 & 99 & \textbf{0.97} & 0.91 & 0.67 \\
 & 2 & 20 & 100 & 0.96 & \textbf{0.94} & 0.87 \\
 & 3 & 14 & 98 & 0.95 & \textbf{0.94} & \textbf{0.90} \\
 & 4 & 11 & 99 & 0.95 & 0.93 & \textbf{0.90} \\
 & 5 & 9 &  99 & 0.95 & 0.93 & \textbf{0.90} \\

\bottomrule
\end{tabular}
\end{sc}
\end{center}
\vskip -0.1in
\end{table}

\begin{table}[t!]
\caption{Accuracy on the \dsb{} dataset.}
\label{tb:sb_comp}
\scriptsize
\vskip 0.15in
\begin{center}
\begin{sc}
\begin{tabular}{lcccrccc}
\toprule
\multirow{2}{*}{\textnormal{model}} & \multicolumn{3}{c}{\textnormal{parameters}} & \multicolumn{3}{c}{\textnormal{attacker}} \\
 & $b$ & $r$ & $|\forest|$ & $A_0$ & $A_1$ & $A_2$ \\
\midrule
\rf  & & & 100 & {\bf 0.96} & 0.58 & 0.18 \\
\rsm & & & 100 & 0.90 & 0.82 & 0.70\\
\midrule
 \multirow{5}{*}{\algoname} 
 & 1 & 33 & 99 & 0.93 & 0.81 & 0.34 \\
 & 2 & 20 & 100 & 0.91 & {\bf 0.83} & 0.72 \\
 & 3 & 14 & 98 & 0.90  & {\bf 0.83} & 0.75 \\
 & 4 & 11 & 99 & 0.88  & {\bf 0.83} & {\bf 0.77} \\
 & 5 & 9 &  99 & 0.86  & {\bf 0.83} & {\bf 0.77} \\
\bottomrule
\end{tabular}
\end{sc}
\end{center}
\vskip -0.1in
\end{table}

\begin{table}[t!]
\caption{Accuracy on the \dwn{} dataset.}
\label{tb:wn_comp}
\scriptsize
\vskip 0.15in
\begin{center}
\begin{sc}
\begin{tabular}{lcccrccc}
\toprule
\multirow{2}{*}{\textnormal{model}} & \multicolumn{3}{c}{\textnormal{parameters}} & \multicolumn{3}{c}{\textnormal{attacker}} \\
 & $b$ & $r$ & $|\forest|$ & $A_0$ & $A_1$ & $A_2$ \\
\midrule
 \rf & & & 300 & \textbf{0.98} & 0.87 & 0.15\\
 \rsm & & & 300 & 0.97 & \textbf{0.92} & 0.71 \\
\midrule
\multirow{5}{*}{\algoname}  
 & 1 & 100 &300 &\textbf{0.98} & 0.89 &0.51 \\
 & 2 & 60 &300 &\textbf{0.98} &\textbf{0.92} & 0.71 \\
 & 3 & 42 &294 &0.97 &\textbf{0.92} &0.74 \\
 & 4 & 33 &297 &0.97 &\textbf{0.92} &0.75 \\
 & 5 & 27 &297 &0.95 &\textbf{0.92} &\textbf{0.79} \\
\bottomrule
\end{tabular}
\end{sc}
\end{center}
\vskip -0.1in
\end{table}

In Tables~\ref{tb:bc_comp},\ref{tb:sb_comp},\ref{tb:wn_comp}
we report the accuracy of \algoname, \rf{} and \rsm{} 
against an attacker that can modify 0, 1 or 2 features.
Indeed, we report the case of no attacks for a more complete evaluation,
but in an adversarial scenario the attacker has no reason for not
conducting an attack.
For \algoname{} we evaluate its robustness
on varying the number of rounds $r$ and the defense strength $b$,
still keeping the total number of trees about constant and comparable
with the size of the forests generated by competitor algorithms.

In all datasets, \rf{} performs best or second best in absence of attacks,
but its accuracy drops significantly under attack. The perturbation of
one single feature is sufficient to harm the model, with a loss of about 40 points in accuracy on \dsb{}, and when attacking two features the accuracy drops under
20\% for \dsb{} and \dwn{} datasets. 
As previously discussed, the $L_0$-norm attack we are
tackling in this work is indeed very powerful and sufficient to fool a very
accurate and effective random forest model.

The \rsm{} model provides good performance in absence of attacks,
meaning that the dataset projection is not disadvantageous, 
and it is much more robust than \rf{} in present of attacks.
However, when attacking two features, \rsm{} exhibits a drop
of 10 to 20 points in accuracy.

The proposed \algoname{} algorithm can provide the best robustness in all attack scenarios. When increasing the defensive $b$, the accuracy slightly decreases in absence of attacks, but always increases under attack, suggesting that large $b$
can be useful even with weaker attacks. 
The performance of \algoname{}
is similar to that of \rsm{} when only one feature is attacked,
but when two features are attacked \algoname{} shows significantly better performance
than \rsm{} with a 10\% relative improvement on both \dsb{} and \dwn{} datasets.

We conclude that \algoname{} is able to outperform state-of-the-art competitors
especially with a stronger attacker.

\subsection{Sensitivity analysis}

In Table~\ref{tb:bc_ra_rpf} we evaluate the sensitivity of \algoname{} w.r.t.\ the number of rounds $r$ on the \dbc{} dataset. Similar results were observed for the other datasets.
As expected, the ensembling strategy improves the accuracy of the resulting model,
and accuracy increases when increasing the number of rounds until a plateau is reached.
We can conclude that using a large number of rounds $r$ improves
the robustness of the trained model.

The impact of $b$ is evaluated in Table~\ref{tb:bc_ba_rpf}.
In this case a trade-off is apparent.
Even if increasing $b$ is expected to increase robustness against stronger attacks,
at the same time it reduces the number of features that can be exploited
when training a single tree. This harms the performance of the whole ensemble
especially with larger values of $b$ or when the dataset contains
a limited number of informative features. The results in Table~\ref{tb:bc_ba_rpf}
show that when using a limited number of trees, accuracy increases with $b$.
But when the forest is sufficiently large to exploit the ensembling benefits,
then the limited accuracy of singles trees plays an important role
making not rewarding the use of larger values of $b$.
For instance, this is the case of the \dbc{} dataset, where we identified only 15 informative features which are difficult to partition in $2b+1$ sets for $b$=5.

We conclude that, while it is beneficial to increase the number of rounds $r$,
it is not always a good strategy to increase $b$, unless the dataset we have at hand
has a sufficiently large set of informative features compared to the attacker strength.

\begin{figure}[t!]
\centering

\begin{tabular}{cc}
\begin{subfigure}
    \centering
    \includegraphics[width=.2\textwidth]{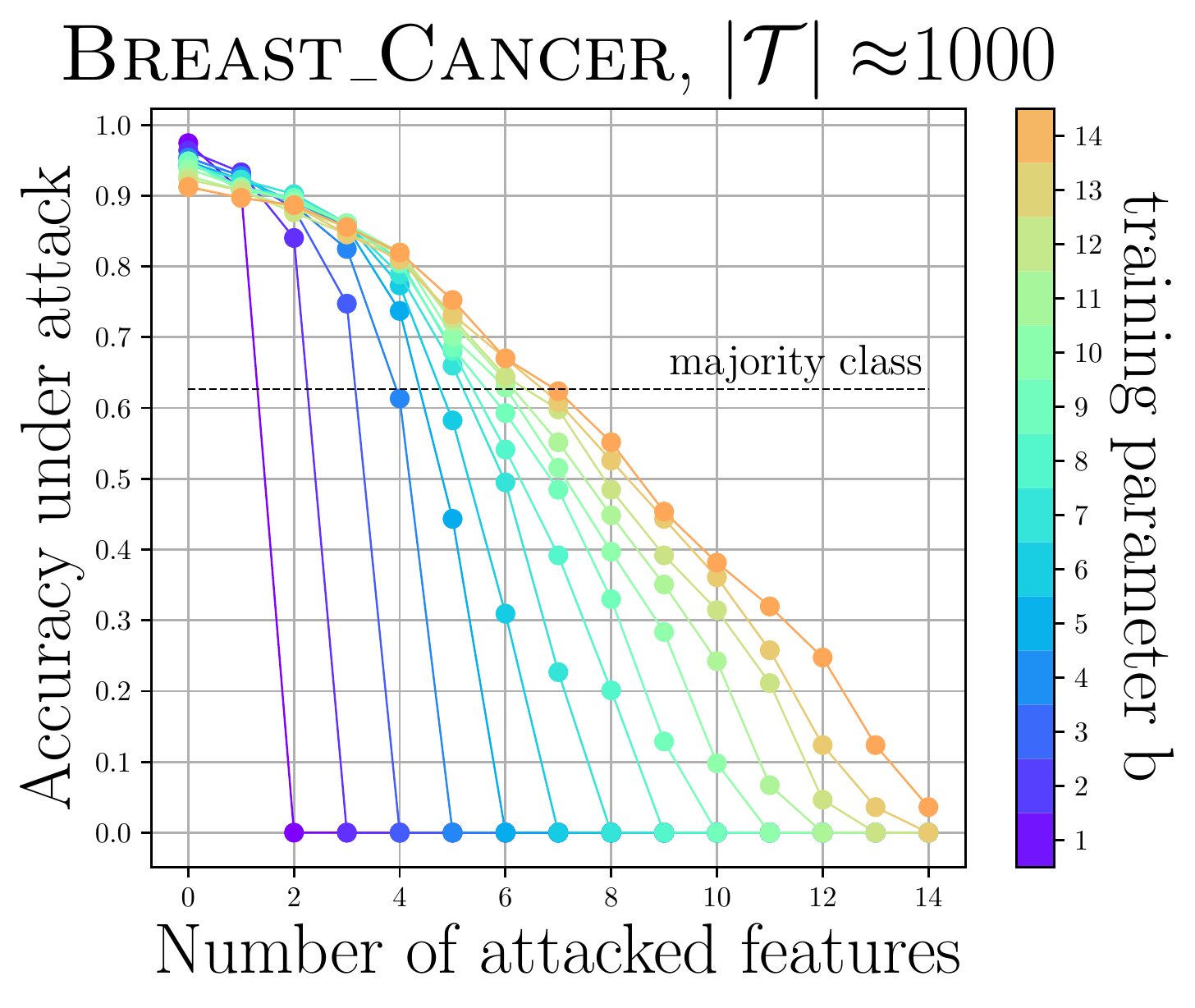}
    \label{fig:bc_flb}
\end{subfigure}
&
\begin{subfigure}
    \centering 
    \includegraphics[width=.2\textwidth]{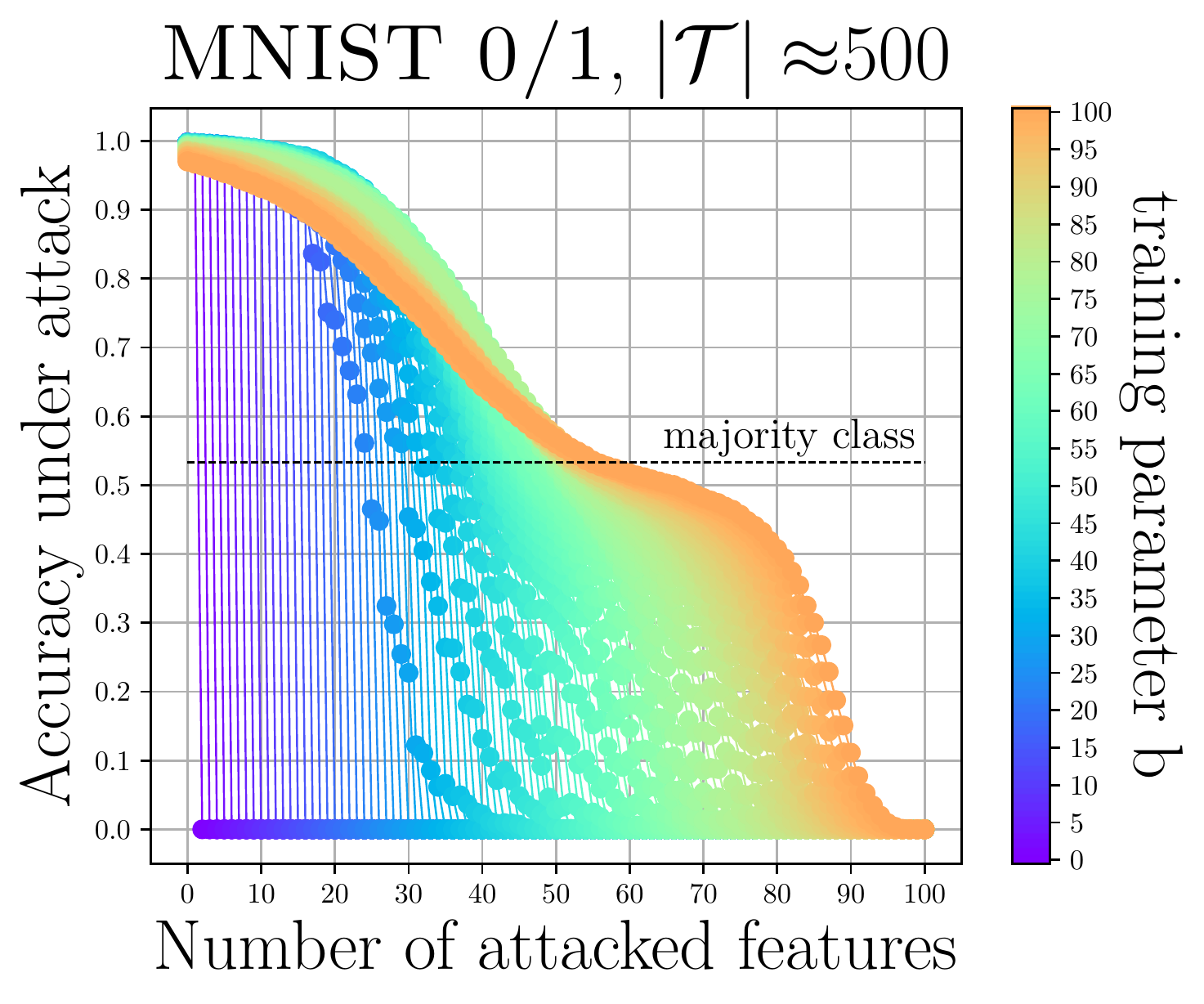}
    \label{fig:m01_flb}
\end{subfigure}
\\

\begin{subfigure}
    \centering 
    \includegraphics[width=.2\textwidth]{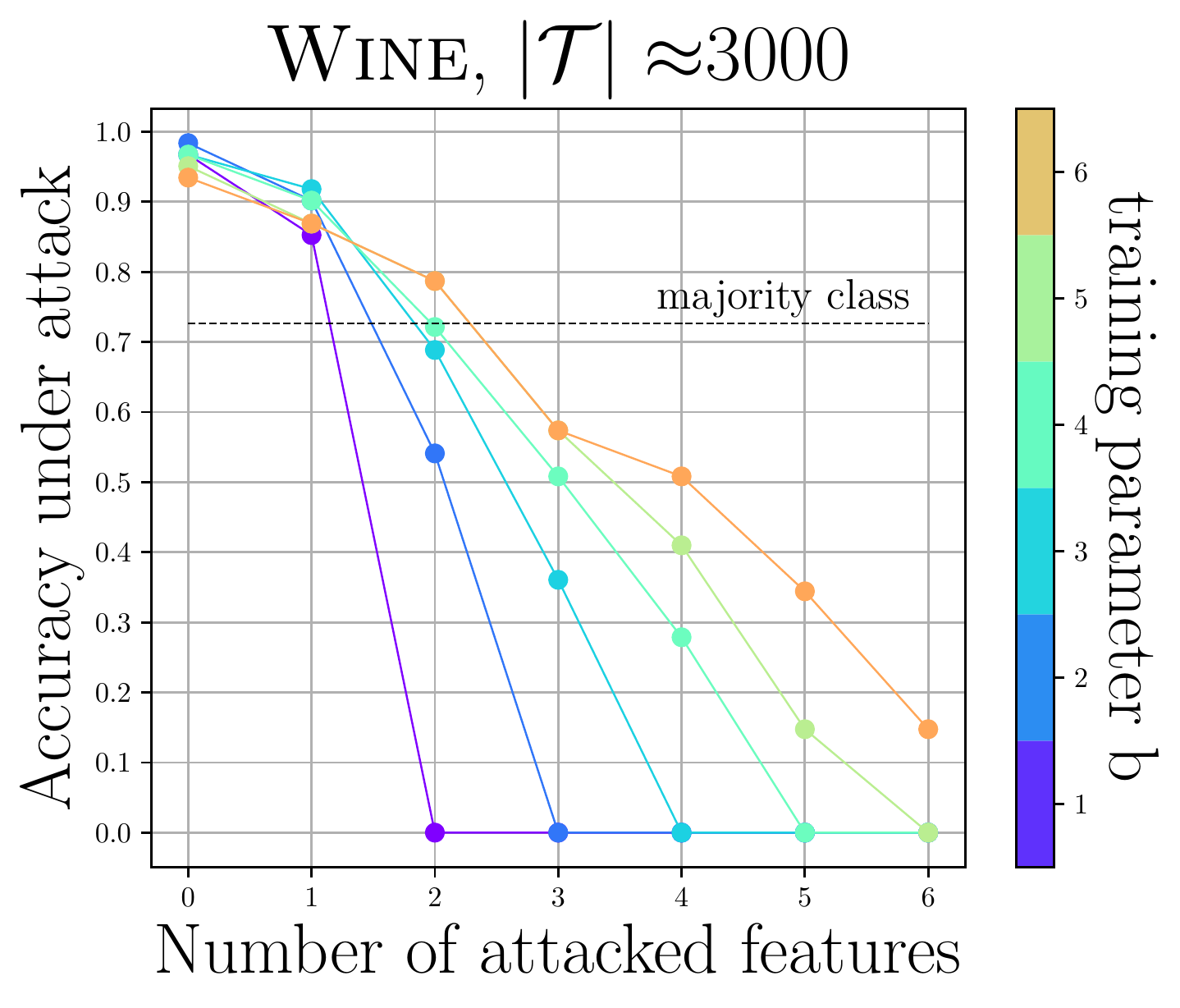}
    \label{fig:bw_flb}
\end{subfigure}
&
\begin{subfigure}
    \centering 
    \includegraphics[width=.2\textwidth]{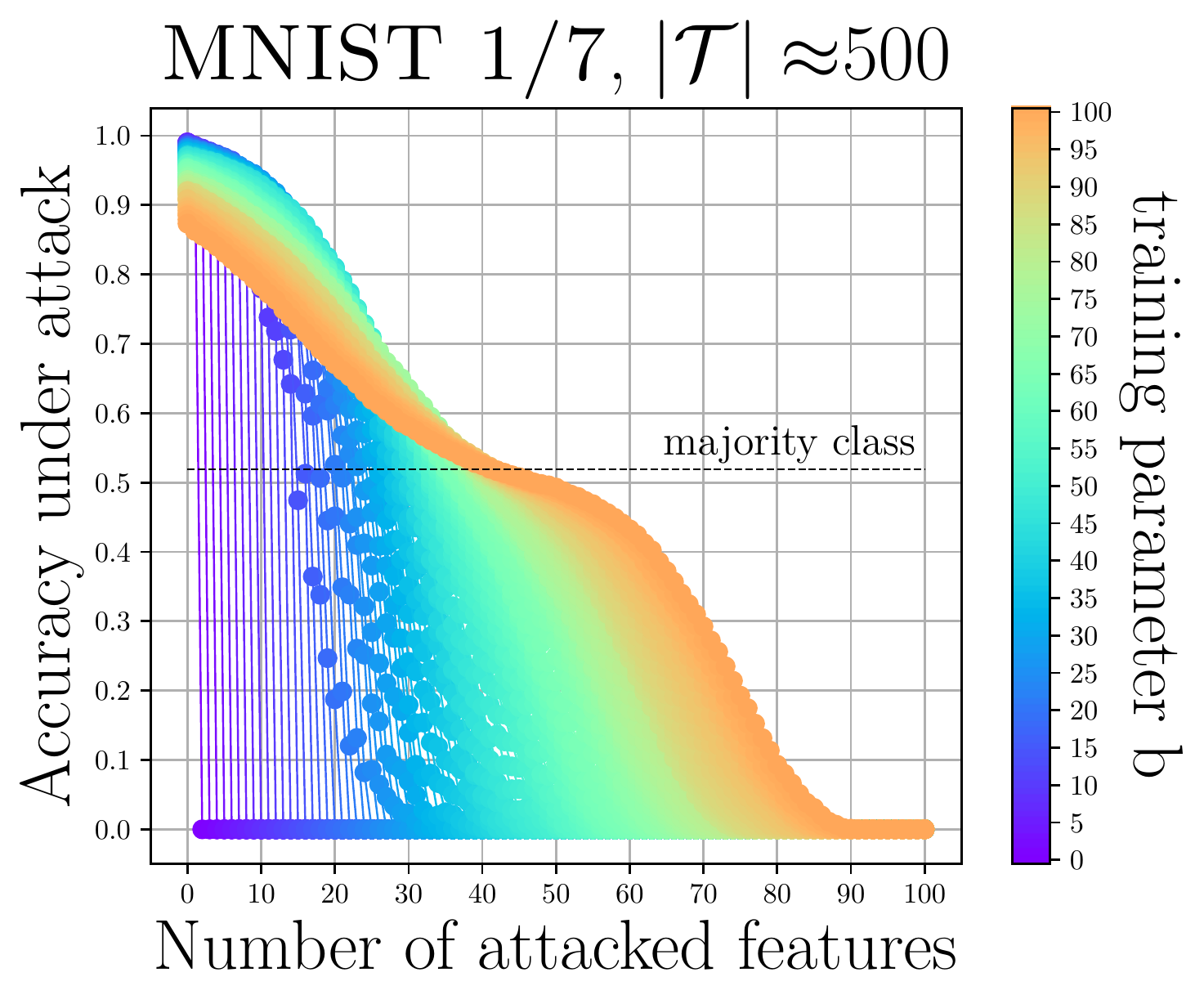}
\end{subfigure}
\\

\begin{subfigure}
    \centering 
    \includegraphics[width=.2\textwidth]{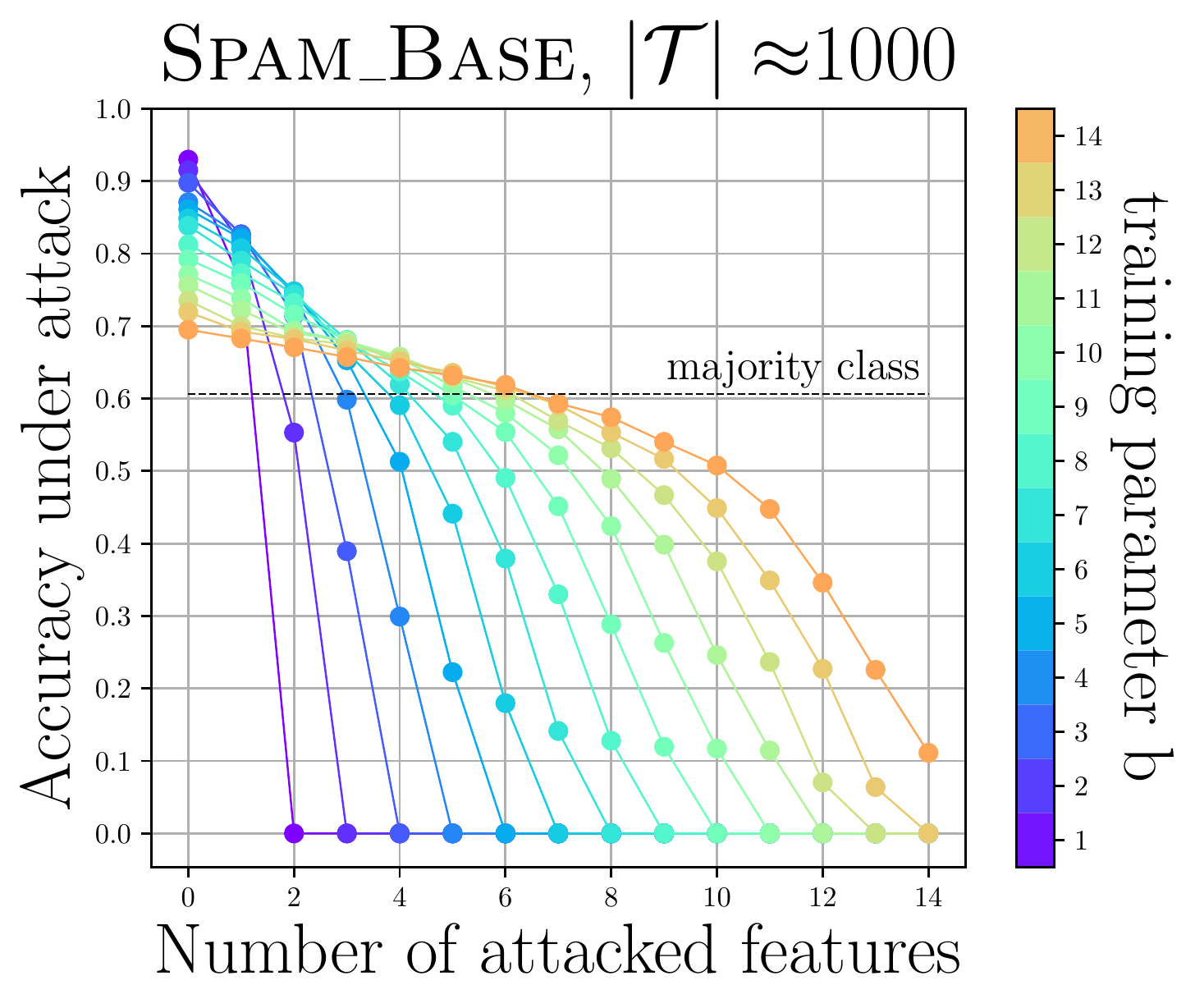}
    \label{fig:sb_flb}
\end{subfigure}

&
\begin{subfigure}
    \centering 
    \includegraphics[width=.2\textwidth]{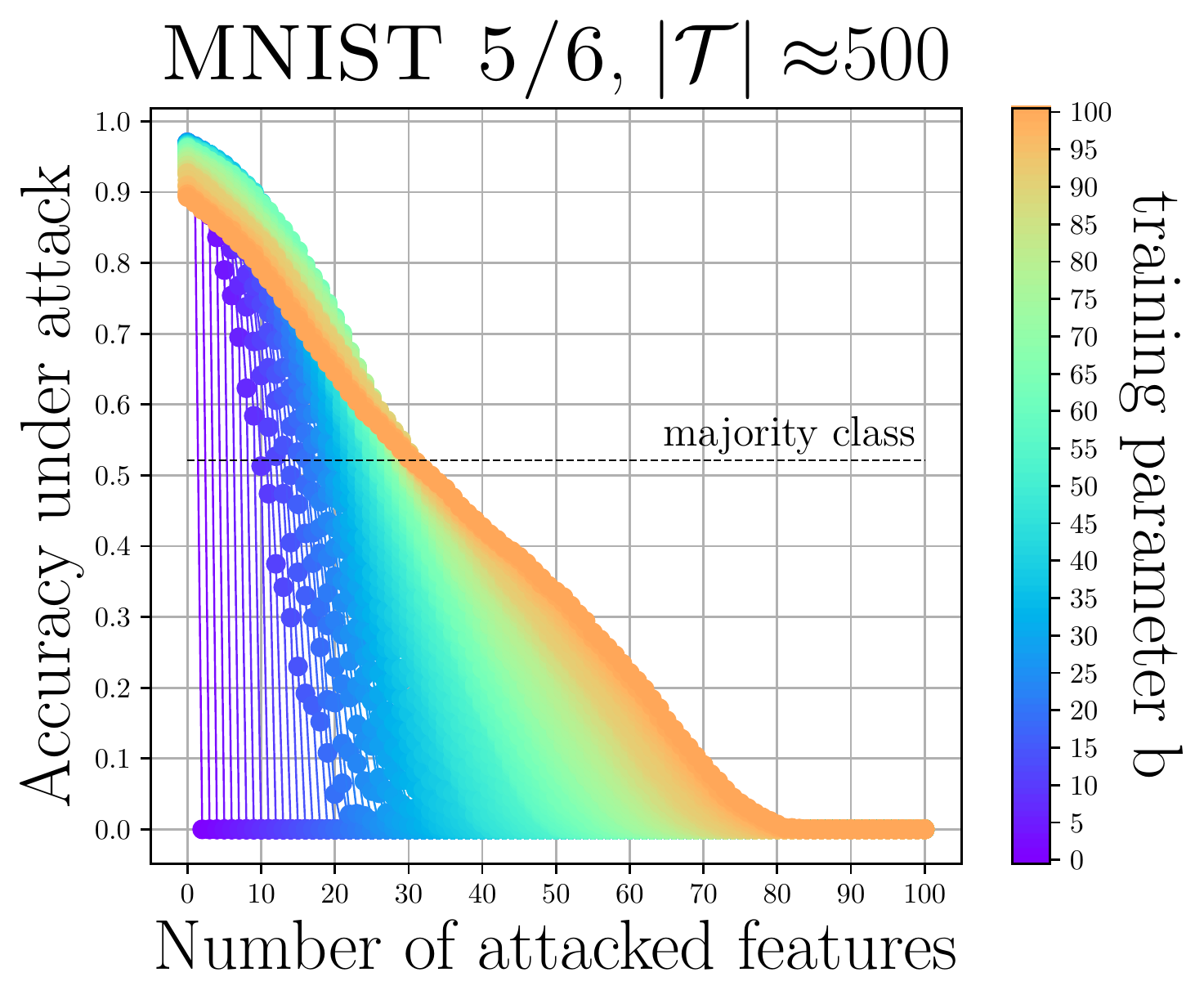}
    \label{fig:m56_flb}
\end{subfigure}
\end{tabular}

\caption{Accuracy lower bounds for large $b$.}
\label{fig:flb_an}
\end{figure}

\begin{table}[t!]
\caption{Accuracy of \algoname{} on the \dbc{} dataset when varying the number of rounds $r$.}
\label{tb:bc_ra_rpf}
\scriptsize
\vskip 0.15in
\begin{center}
\begin{sc}
\begin{tabular}{ccccc}
\toprule
		\multirow{2}{*}{$b$} & \multirow{2}{*}{$A_k$} & \multicolumn{3}{c}{number of rounds $r$} \\
		 & & 1 & 15 & 30 \\
		\midrule
		\multirow{3}{*}{1}
		 & 0 & 0.964 & 0.969 & \textbf{0.974} \\
		 & 1 & 0.881 & \textbf{0.902} & \textbf{0.902} \\
		 & 2 & 0.108 & 0.670 & \textbf{0.691} \\\midrule

		\multirow{3}{*}{2}
		 & 0 & 0.954 & \textbf{0.964} & \textbf{0.964} \\
		 & 1 & 0.902 & \textbf{0.943} & \textbf{0.943} \\
		 & 2 & 0.804 & 0.866 & \textbf{0.871} \\ \midrule

		\multirow{3}{*}{3}
		 & 0 & \textbf{0.954} & \textbf{0.954} & \textbf{0.954} \\
		 & 1 & 0.923 & 0.938 & \textbf{0.938} \\
		 & 2 & 0.851 & 0.897 & \textbf{0.902} \\ \midrule

		\multirow{3}{*}{4}
		 & 0 & 0.943 & \textbf{0.954} & \textbf{0.954} \\
		 & 1 & 0.907 & 0.928 & \textbf{0.933} \\
		 & 2 & 0.871 & 0.887 & \textbf{0.902} \\ \midrule

		\multirow{3}{*}{5}
		 & 0 & 0.933 & \textbf{0.954} & 0.948 \\
		 & 1 & 0.912 & 0.923 & \textbf{0.928} \\
		 & 2 & 0.876 & 0.892 & \textbf{0.897} \\
\bottomrule
\end{tabular}
\end{sc}
\end{center}
\vskip -0.1in
\end{table}

\begin{table}[t!]
\caption{Accuracy of \algoname{} on the \dbc{} dataset when varying the partitioning parameter $b$.}
\label{tb:bc_ba_rpf}
\scriptsize
\vskip 0.15in
\begin{center}
\begin{sc}
\begin{tabular}{ccccccc}
\toprule
		\multirow{2}{*}{$|\forest|$} & \multirow{2}{*}{$A_k$} & \multicolumn{5}{c}{parameter $b$} \\
		 & & 1 & 2 & 3 & 4 & 5 \\
		\midrule
		\multirow{3}{*}{50} & 0 & \textbf{0.969} & \textbf{0.969} & 0.964 & 0.954 & 0.948 \\
		 & 1 & 0.902 & 0.938 & 0.938 & \textbf{0.943} & 0.928 \\
		 & 2 & 0.628 & 0.871 & 0.876 & \textbf{0.892} & \textbf{0.892} \\ \midrule

		\multirow{3}{*}{75} 
		 & 0 & \textbf{0.969} & 0.964 & 0.959 & 0.959 & 0.943 \\
		 & 1 & 0.907 & \textbf{0.943} & \textbf{0.943} & 0.933 & 0.928 \\
		 & 2 & 0.629 & 0.866 & 0.902 & \textbf{0.902} & 0.897 \\ \midrule

		\multirow{3}{*}{100} & 0 & \textbf{0.974} & 0.964 & 0.954 & 0.954 & 0.948 \\
		 & 1 & 0.912 & \textbf{0.943} & 0.938 & 0.933 & 0.928 \\
		 & 2 & 0.675 & 0.871 & \textbf{0.902} & 0.897 & 0.897 \\
\bottomrule
\end{tabular}
\end{sc}
\end{center}
\vskip -0.1in
\end{table}

\subsection{Lower Bound Analysis}

In table \ref{tb:bc_lb} we report the accuracy under attack computed with {\sc brute\_force}, {\sc elb} and {\sc flb} with respect to $A_1$ and $A_2$.
With attacks on 1 feature {\sc elb} and {\sc flb} have the same estimate of the accuracy under attack. With attacks on 2 features {\sc elb} performs better with small $b$. In general, both models predict an accuracy very close to the real calculated one. This means that, the proposed lower bounds can certify the non attackability of a large portion of instances without the cost of the brute-force exploration.

\begin{table}[t]
\caption{Lower bound analysis \dbc{}. \algoname{} with $|\mathcal{T}| \approx 100$}
\label{tb:bc_lb}
\scriptsize
\vskip 0.15in
\begin{center}
\begin{sc}
\begin{tabular}{cllllllcccc}
\toprule
\multicolumn{1}{c}{} & \multicolumn{3}{c}{$A_1$} & \multicolumn{3}{c}{$A_2$} \\
$b$ & bf & elb & flb & bf & elb & flb \\ 
\midrule
1 & 0.912 &0.897 &0.897 &0.670 &0.000 &0.000 \\
2 & 0.943 &0.938 &0.938 &0.870 &0.835 &0.820 \\
3 & 0.938 &0.938 &0.938 &0.900 &0.881 &0.876 \\
4 & 0.933 &0.933 &0.933 &0.900 &0.887 &0.887 \\
5 & 0.928 &0.928 &0.928 &0.900 &0.892 &0.892 \\
\bottomrule
\end{tabular}
\end{sc}
\end{center}
\vskip -0.1in
\end{table}

In Figure~\ref{fig:flb_an} we show the accuracy lower bound computed on different dataset and on varying $b$ and the attacker $A$.
The computational efficiency of the proposed bound allows to compute
the minimum accuracy for large values of $b$ and large attacker budgets.
The figure shows how larger values of $b$ allow to sustain a larger attacker strength.
Of course, when the attacker becomes too strong compared with the number
of relevant features in the dataset, then the accuracy of \algoname{} drops.
We include in our analysis three dataset generated from MINST by isolating instances of two digits. The lower bounds allows us to state that a reasonable accuracy can be achieved also when attacking more than 20 features. The weakest dataset
is that of digits 5 vs.\ 6, where clearly the attacker requires to change fewer
pixels to generate a misclassification.


\section{Conclusion}
\label{sec:conclusion}

This paper proposes \algoname, a new algorithm to generate forests of decision trees, based on  random equi-partitioning of the feature set, along with a projections of the dataset on these partitions before training each single decision tree.
The method is proven to be resilient against evasion attacks, and, more importantly, we are able to certificate in a very efficient way that, given a test dataset, some of the instances cannot be attacked at all, thus avoiding the
costly computation of all the possible evasion attacks.

The experimental evaluation, carried out on publicly available
datasets, is promising and ouperforms the main direct competitor, based on ensembles build on random sampling of the features.
Moreover, we also show that our certified lower bounds on the accuracy under attack are a very close approximation of the actual accuracy.


\bibliography{paper}

\begin{thebibliography}{29}
\providecommand{\natexlab}[1]{#1}
\providecommand{\url}[1]{\texttt{#1}}
\expandafter\ifx\csname urlstyle\endcsname\relax
  \providecommand{\doi}[1]{doi: #1}\else
  \providecommand{\doi}{doi: \begingroup \urlstyle{rm}\Url}\fi

\bibitem[Biggio \& Roli(2018)Biggio and Roli]{BiggioR17}
Biggio, B. and Roli, F.
\newblock Wild patterns: Ten years after the rise of adversarial machine
  learning.
\newblock \emph{Pattern Recognition}, 84:\penalty0 317--331, 2018.

\bibitem[Biggio et~al.(2010)Biggio, Fumera, and Roli]{biggio2010multiple}
Biggio, B., Fumera, G., and Roli, F.
\newblock Multiple classifier systems for robust classifier design in
  adversarial environments.
\newblock \emph{International Journal of Machine Learning and Cybernetics},
  1\penalty0 (1-4):\penalty0 27--41, 2010.

\bibitem[Biggio et~al.(2011)Biggio, Nelson, and Laskov]{BiggioNL11}
Biggio, B., Nelson, B., and Laskov, P.
\newblock Support vector machines under adversarial label noise.
\newblock In \emph{{ACML}}, pp.\  97--112, 2011.

\bibitem[Biggio et~al.(2013)Biggio, Corona, Maiorca, Nelson, Srndic, Laskov,
  Giacinto, and Roli]{BiggioCMNSLGR13}
Biggio, B., Corona, I., Maiorca, D., Nelson, B., Srndic, N., Laskov, P.,
  Giacinto, G., and Roli, F.
\newblock Evasion attacks against machine learning at test time.
\newblock In \emph{{ECML} {PKDD}}, pp.\  387--402, 2013.

\bibitem[Biggio et~al.(2014)Biggio, Fumera, and Roli]{BiggioFR14}
Biggio, B., Fumera, G., and Roli, F.
\newblock Security evaluation of pattern classifiers under attack.
\newblock \emph{{IEEE} Trans. Knowl. Data Eng.}, 26\penalty0 (4):\penalty0
  984--996, 2014.

\bibitem[Breiman(2001)]{breiman2001randomforest}
Breiman, L.
\newblock Random forests.
\newblock \emph{Machine Learning}, 45\penalty0 (1):\penalty0 5--32, 2001.

\bibitem[Carlini \& Wagner(2017)Carlini and Wagner]{Carlini017}
Carlini, N. and Wagner, D.~A.
\newblock Towards evaluating the robustness of neural networks.
\newblock In \emph{S\&P}, pp.\  39--57, 2017.

\bibitem[Chen et~al.(2019{\natexlab{a}})Chen, Zhang, Boning, and
  Hsieh]{ChenZBH19}
Chen, H., Zhang, H., Boning, D.~S., and Hsieh, C.
\newblock Robust decision trees against adversarial examples.
\newblock In \emph{{ICML}}, pp.\  1122--1131, 2019{\natexlab{a}}.

\bibitem[Chen et~al.(2019{\natexlab{b}})Chen, Zhang, Si, Li, Boning, and
  Hsieh]{chen2019robustness}
Chen, H., Zhang, H., Si, S., Li, Y., Boning, D., and Hsieh, C.-J.
\newblock Robustness verification of tree-based models.
\newblock In \emph{Advances in Neural Information Processing Systems}, pp.\
  12317--12328, 2019{\natexlab{b}}.

\bibitem[Chollet(2017)]{Chollet:2017:DLP:3203489}
Chollet, F.
\newblock \emph{Deep Learning with Python}.
\newblock Manning Publications Co., Greenwich, CT, USA, 1st edition, 2017.
\newblock ISBN 1617294438, 9781617294433.

\bibitem[Dang et~al.(2017)Dang, Huang, and Chang]{DangHC17}
Dang, H., Huang, Y., and Chang, E.
\newblock Evading classifiers by morphing in the dark.
\newblock In \emph{{CCS}}, pp.\  119--133, 2017.

\bibitem[Goodfellow et~al.(2015)Goodfellow, Shlens, and
  Szegedy]{GoodfellowSS15}
Goodfellow, I.~J., Shlens, J., and Szegedy, C.
\newblock Explaining and harnessing adversarial examples.
\newblock In \emph{ICLR}, 2015.

\bibitem[Gu \& Rigazio(2015)Gu and Rigazio]{GuR14}
Gu, S. and Rigazio, L.
\newblock Towards deep neural network architectures robust to adversarial
  examples.
\newblock In \emph{{ICLR}, Workshop Track Proceedings}, 2015.

\bibitem[Ho(1998)]{ho1998random}
Ho, T.~K.
\newblock The random subspace method for constructing decision forests.
\newblock \emph{IEEE transactions on pattern analysis and machine
  intelligence}, 20\penalty0 (8):\penalty0 832--844, 1998.

\bibitem[Huang et~al.(2011)Huang, Joseph, Nelson, Rubinstein, and
  Tygar]{HuangJNRT11}
Huang, L., Joseph, A.~D., Nelson, B., Rubinstein, B. I.~P., and Tygar, J.~D.
\newblock Adversarial machine learning.
\newblock In \emph{AISec}, pp.\  43--58, 2011.

\bibitem[Kantchelian et~al.(2016{\natexlab{a}})Kantchelian, Tygar, and
  Joseph]{kantchelian2016evasion}
Kantchelian, A., Tygar, J.~D., and Joseph, A.
\newblock Evasion and hardening of tree ensemble classifiers.
\newblock In \emph{International Conference on Machine Learning}, pp.\
  2387--2396, 2016{\natexlab{a}}.

\bibitem[Kantchelian et~al.(2016{\natexlab{b}})Kantchelian, Tygar, and
  Joseph]{KantchelianTJ16}
Kantchelian, A., Tygar, J.~D., and Joseph, A.~D.
\newblock Evasion and hardening of tree ensemble classifiers.
\newblock In \emph{ICML}, pp.\  2387--2396, 2016{\natexlab{b}}.

\bibitem[Lowd \& Meek(2005)Lowd and Meek]{LowdM05}
Lowd, D. and Meek, C.
\newblock Adversarial learning.
\newblock In \emph{{SIGKDD}}, pp.\  641--647, 2005.

\bibitem[Moosavi{-}Dezfooli et~al.(2016)Moosavi{-}Dezfooli, Fawzi, and
  Frossard]{Moosavi-Dezfooli16}
Moosavi{-}Dezfooli, S., Fawzi, A., and Frossard, P.
\newblock Deepfool: {A} simple and accurate method to fool deep neural
  networks.
\newblock In \emph{{CVPR}}, pp.\  2574--2582, 2016.

\bibitem[Nelson et~al.(2010)Nelson, Rubinstein, Huang, Joseph, Lau, Lee, Rao,
  Tran, and Tygar]{NelsonRHJLLRTT10}
Nelson, B., Rubinstein, B. I.~P., Huang, L., Joseph, A.~D., Lau, S., Lee,
  S.~J., Rao, S., Tran, A., and Tygar, J.~D.
\newblock Near-optimal evasion of convex-inducing classifiers.
\newblock In \emph{{AISTATS}}, pp.\  549--556, 2010.

\bibitem[Nguyen et~al.(2015)Nguyen, Yosinski, and Clune]{NguyenYC15}
Nguyen, A.~M., Yosinski, J., and Clune, J.
\newblock Deep neural networks are easily fooled: High confidence predictions
  for unrecognizable images.
\newblock In \emph{{CVPR}}, pp.\  427--436, 2015.

\bibitem[Papernot et~al.(2016{\natexlab{a}})Papernot, McDaniel, Jha,
  Fredrikson, Celik, and Swami]{PapernotMJFCS16}
Papernot, N., McDaniel, P.~D., Jha, S., Fredrikson, M., Celik, Z.~B., and
  Swami, A.
\newblock The limitations of deep learning in adversarial settings.
\newblock In \emph{EuroS{\&}P}, pp.\  372--387, 2016{\natexlab{a}}.

\bibitem[Papernot et~al.(2016{\natexlab{b}})Papernot, McDaniel, Wu, Jha, and
  Swami]{PapernotM0JS16}
Papernot, N., McDaniel, P.~D., Wu, X., Jha, S., and Swami, A.
\newblock Distillation as a defense to adversarial perturbations against deep
  neural networks.
\newblock In \emph{S\&P}, pp.\  582--597, 2016{\natexlab{b}}.

\bibitem[Simonyan \& Zisserman(2014)Simonyan and Zisserman]{simonyan2014very}
Simonyan, K. and Zisserman, A.
\newblock Very deep convolutional networks for large-scale image recognition.
\newblock \emph{arXiv preprint arXiv:1409.1556}, 2014.

\bibitem[Srndic \& Laskov(2014)Srndic and Laskov]{SrndicL14}
Srndic, N. and Laskov, P.
\newblock Practical evasion of a learning-based classifier: {A} case study.
\newblock In \emph{S\&P}, pp.\  197--211, 2014.

\bibitem[Su et~al.(2019)Su, Vargas, and Sakurai]{su2019one}
Su, J., Vargas, D.~V., and Sakurai, K.
\newblock One pixel attack for fooling deep neural networks.
\newblock \emph{IEEE Transactions on Evolutionary Computation}, 2019.

\bibitem[Szegedy et~al.(2014)Szegedy, Zaremba, Sutskever, Bruna, Erhan,
  Goodfellow, and Fergus]{SzegedyZSBEGF13}
Szegedy, C., Zaremba, W., Sutskever, I., Bruna, J., Erhan, D., Goodfellow,
  I.~J., and Fergus, R.
\newblock Intriguing properties of neural networks.
\newblock In \emph{{ICLR}}, 2014.

\bibitem[Tolomei et~al.(2017)Tolomei, Silvestri, Haines, and
  Lalmas]{tolomei2017kdd}
Tolomei, G., Silvestri, F., Haines, A., and Lalmas, M.
\newblock Interpretable predictions of tree-based ensembles via actionable
  feature tweaking.
\newblock In \emph{{SIGKDD}}, pp.\  465--474, 2017.

\bibitem[Xiao et~al.(2015)Xiao, Biggio, Nelson, Xiao, Eckert, and
  Roli]{XiaoBNXER15}
Xiao, H., Biggio, B., Nelson, B., Xiao, H., Eckert, C., and Roli, F.
\newblock Support vector machines under adversarial label contamination.
\newblock \emph{Neurocomputing}, 160:\penalty0 53--62, 2015.

\end{thebibliography}
\bibliographystyle{icml2019}

\end{document}